\newcommand{\figA}{\begin{figure}[htb]
\begin{center}
\begin{tabular}{cc}
\includegraphics[width=3.0in]{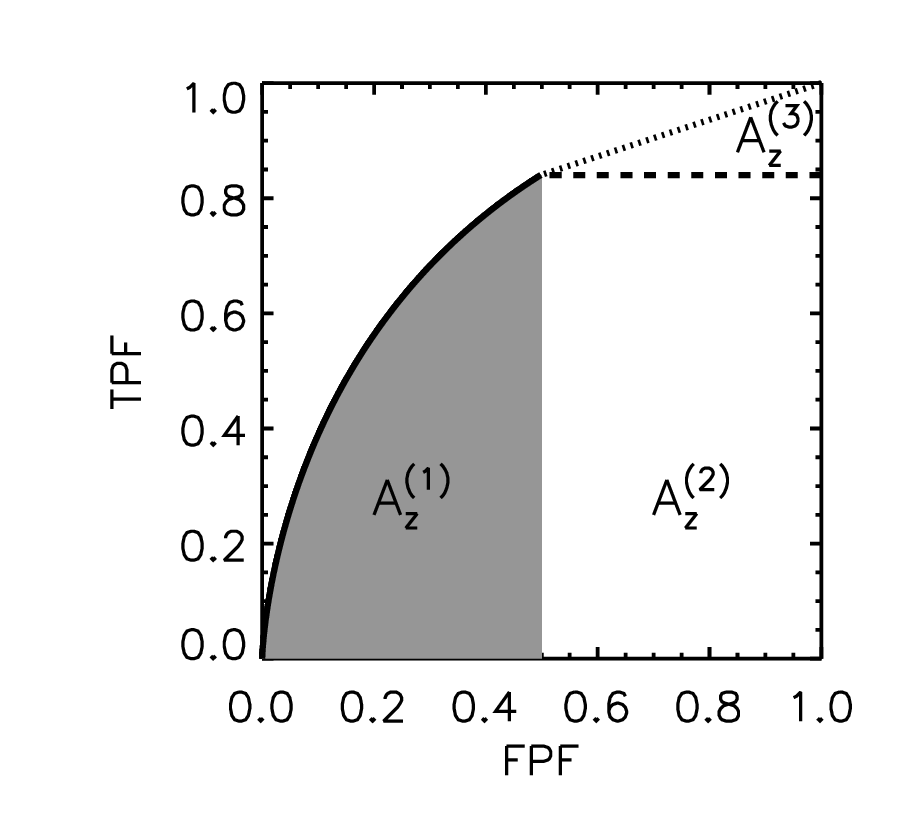} &
\includegraphics[width=3.0in]{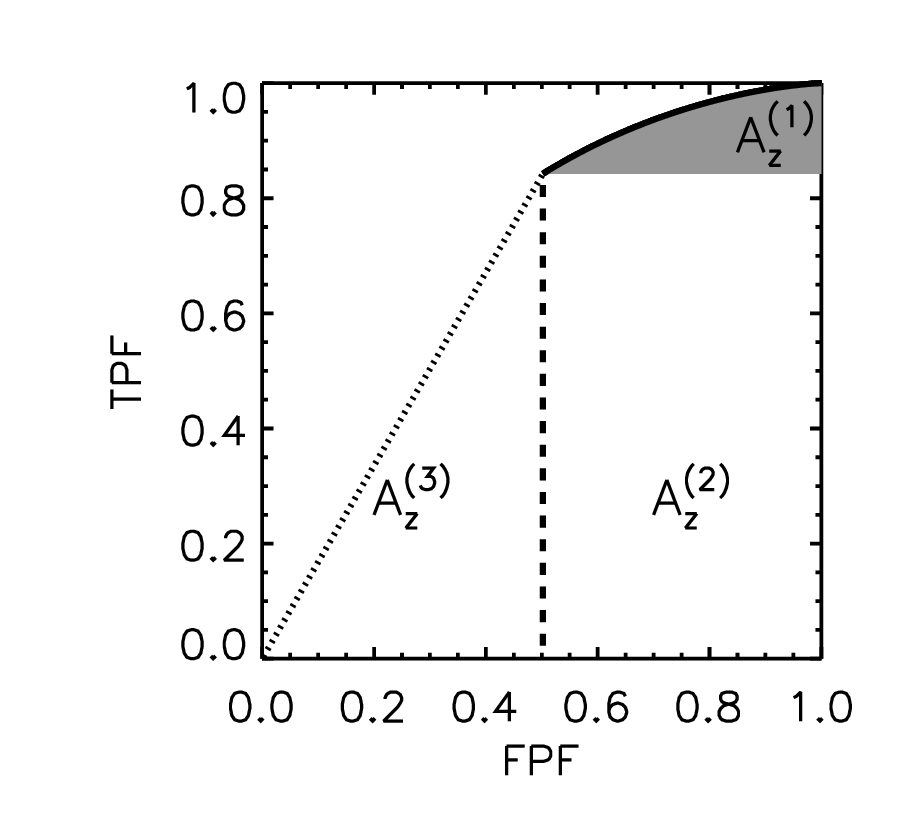}\\
(a) & (b)
\end{tabular}
\end{center}
\caption{\label{figA}Handling unrated ROC cases with the proposed truncation analysis. (a) With lefthand truncation, the partial curve (shown as the solid line) results in the reduced area $A_{z}^{(1)}$ represented by the shaded region. Assigning the minimum rating ($\lambda$ = $-\infty$) to all unrated negative cases provides an additional area $A_{z}^{(2)}$ by completing the curve with the horizontal dashed line. This completion disallows credit for guessing, whereas also assigning the minimum rating to unrated positive cases produces the dotted-line guessing extension with area increment $A_{z}^{(3)}$. (b) ROC curve completion based on righthand truncation. The ROC curves were computed using a single feature.}
\end{figure}}
\newcommand{\figB}{\begin{figure}[htb]
\begin{center}
\includegraphics[width=3.0in]{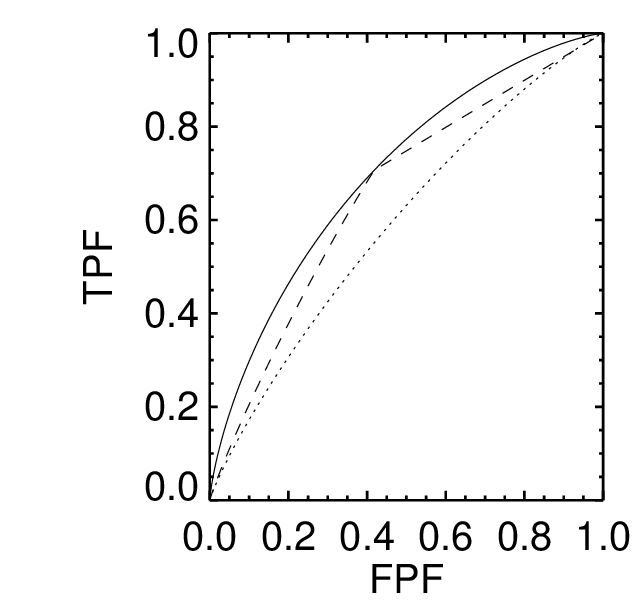} 
\end{center}
\caption{\label{figB}ROC curves from the ideal and truncated ideal observers. A single feature was considered. The dotted and solid lines represent the ideal observer respectively with and without internal noise of variance $\sigma^{2}$. The dashed line is from the thresholded model with the same internal-noise variance.}
\vspace*{-0.1in}
\end{figure}}
\newcommand{\figE}{\begin{figure}[htb]
\begin{tabular}{ccc}
$\sigma_{1}$ = 0.25 & $\sigma_{1}$ = 1.0 & $\sigma_{1}$ = 3.0\\
\includegraphics[width=2.0in]{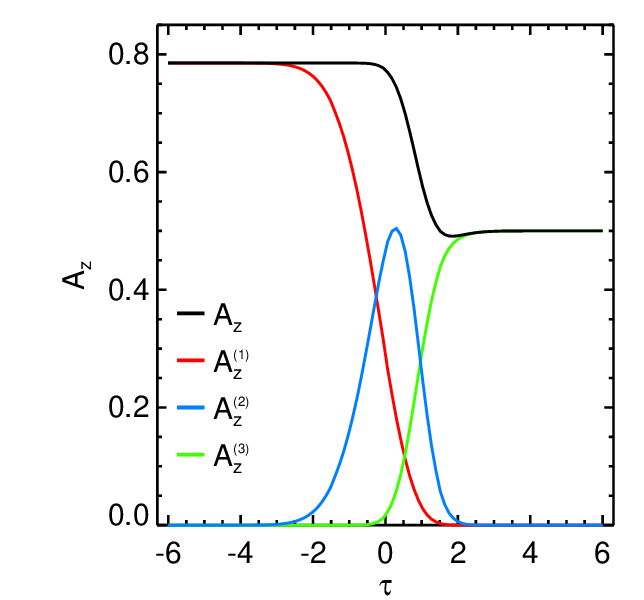} &
\includegraphics[width=2.0in]{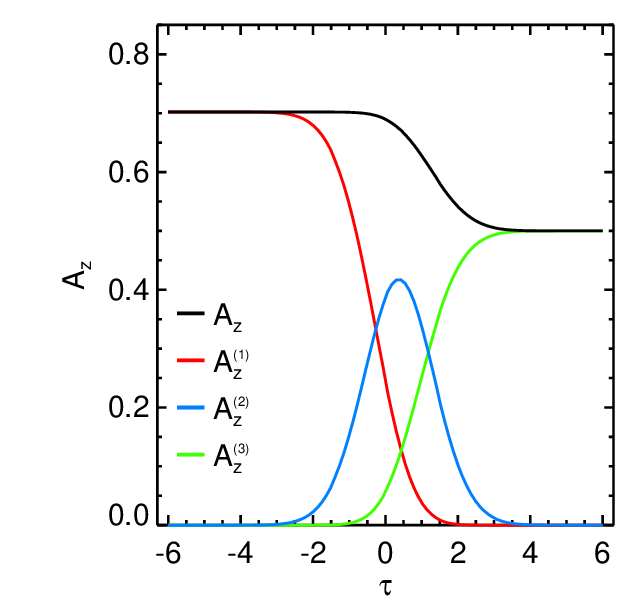} &
\includegraphics[width=2.0in]{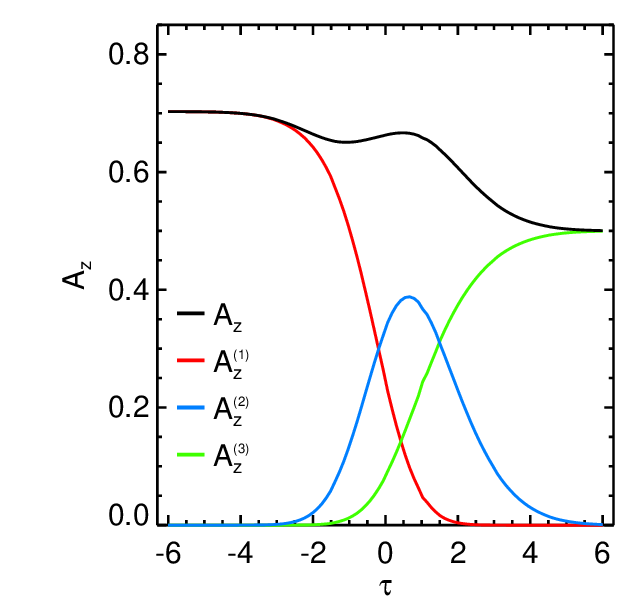}\\
(a) & (b) & (c)\\
\includegraphics[width=2.0in]{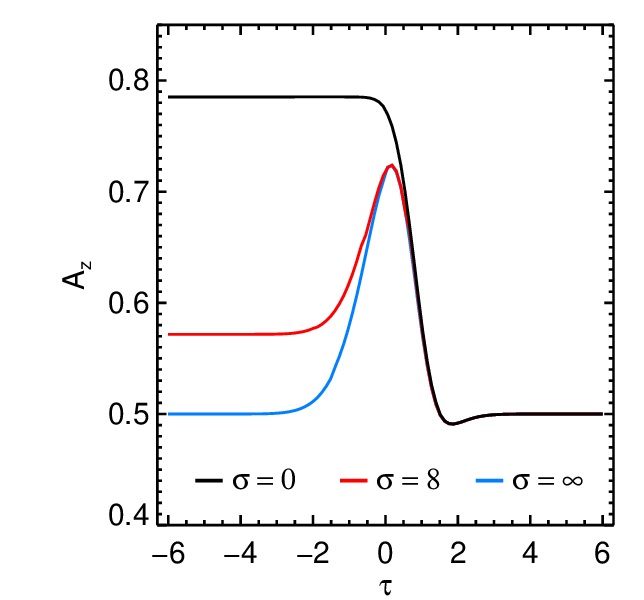} &
\includegraphics[width=2.0in]{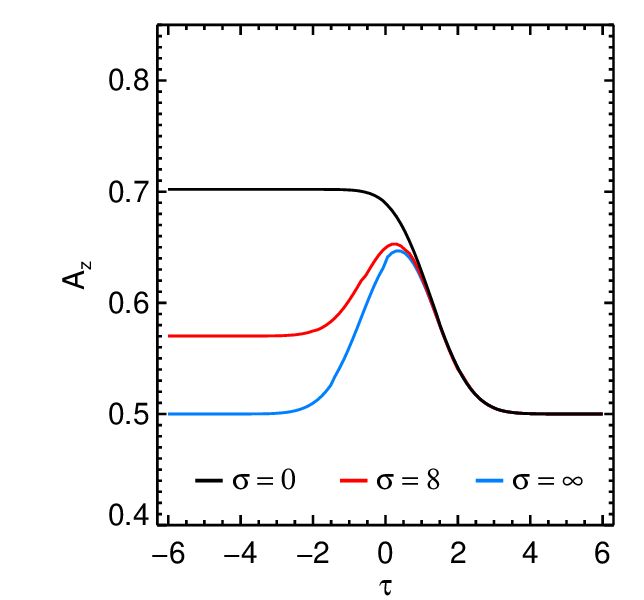} &
\includegraphics[width=2.0in]{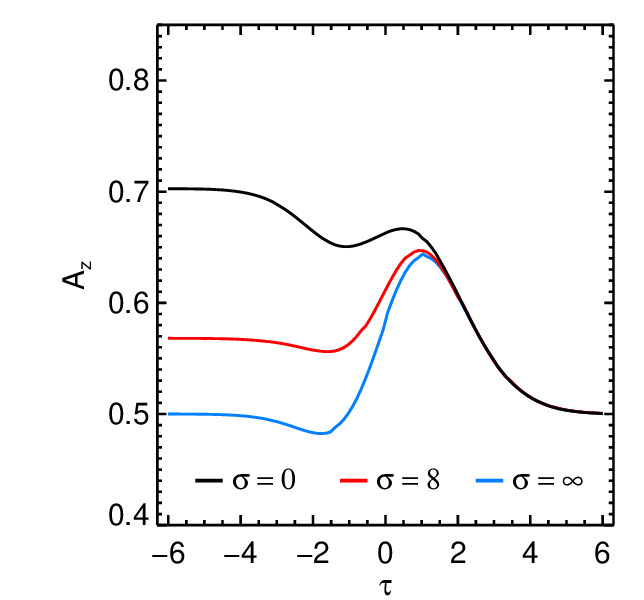}\\
(d) & (e) & (f)\\
\end{tabular}
\caption{\label{figE}Comparison of truncation model performance for three single-feature simulations. In these simulations, the normal feature followed the standard distribution, while the abnormal feature mean was 0.75. From left to right, the standard deviation for the abnormal feature  was 0.25, 1.0 and 3.0. The top row (plots a-c) shows overall performance (in black) without internal noise. Also shown are the analysis (red), gist (blue) and guessing (green) contributions. The bottom row (plots d-f) displays overall performance as a function of threshold for internal noise $\sigma$ $\in$ $\{0,8,\infty\}$.}
\vspace*{-0.1in}
\end{figure}}

\newcommand{\figF}{\begin{figure}[htb]
\begin{center}
\begin{tabular}{ccccc}
& $\sigma$ = 0 & $\sigma$ = 4 & $\sigma$ = 12 & $\sigma$ = $\infty$\\
\raisebox{0.8in}{$\sigma_{1}$ = 0.25} &
\includegraphics[width=1.5in]{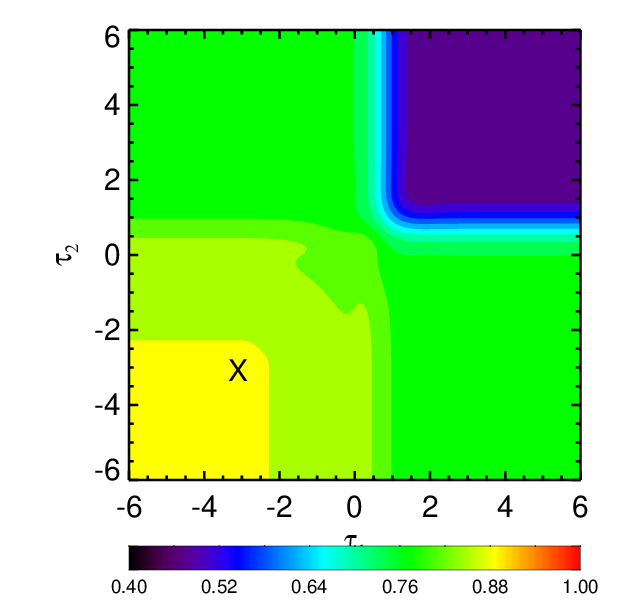} &
\includegraphics[width=1.5in]{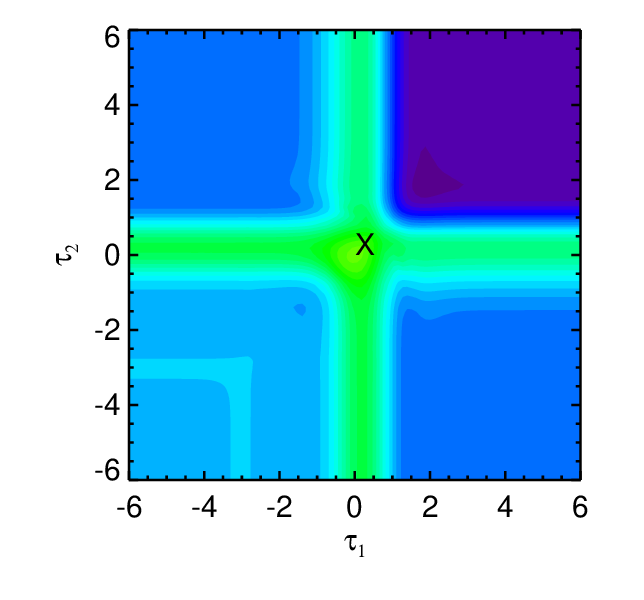} &
\includegraphics[width=1.5in]{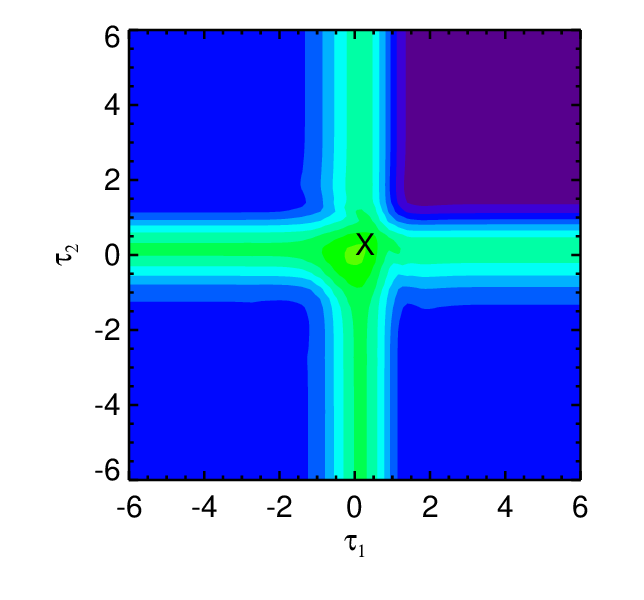} &
\includegraphics[width=1.5in]{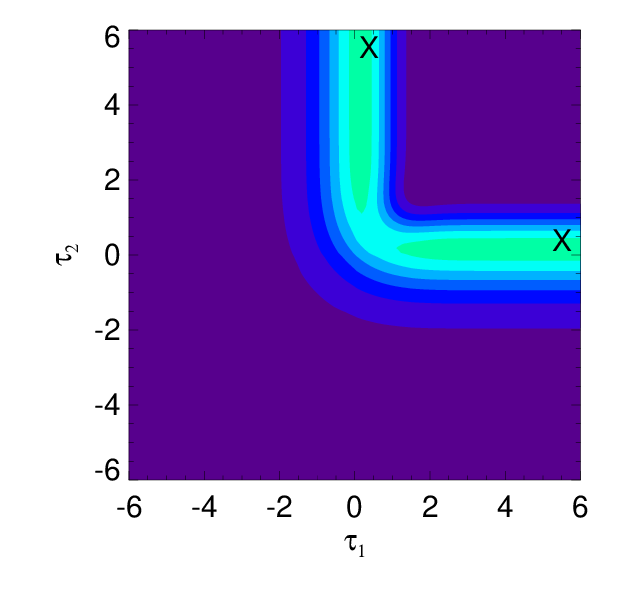}\\
\raisebox{0.8in}{$\sigma_{1}$ = 1.0} &
\includegraphics[width=1.5in]{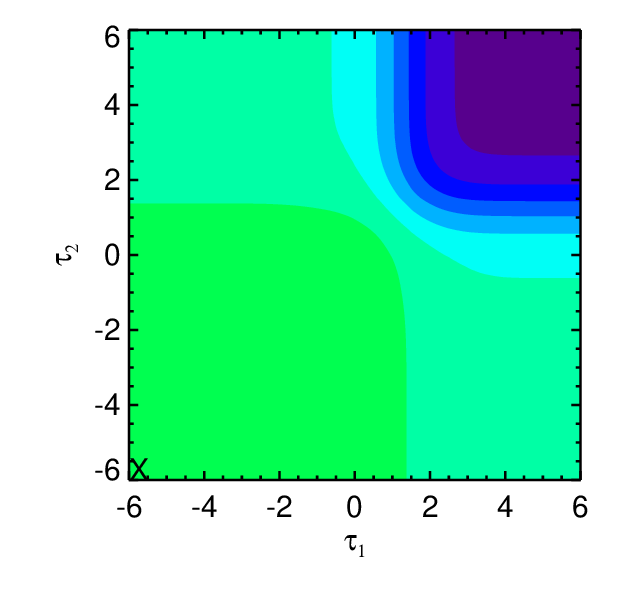} &
\includegraphics[width=1.5in]{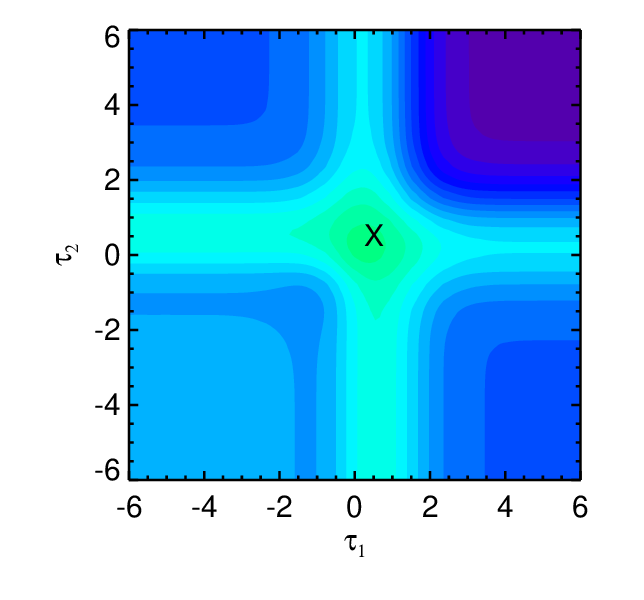} &
\includegraphics[width=1.5in]{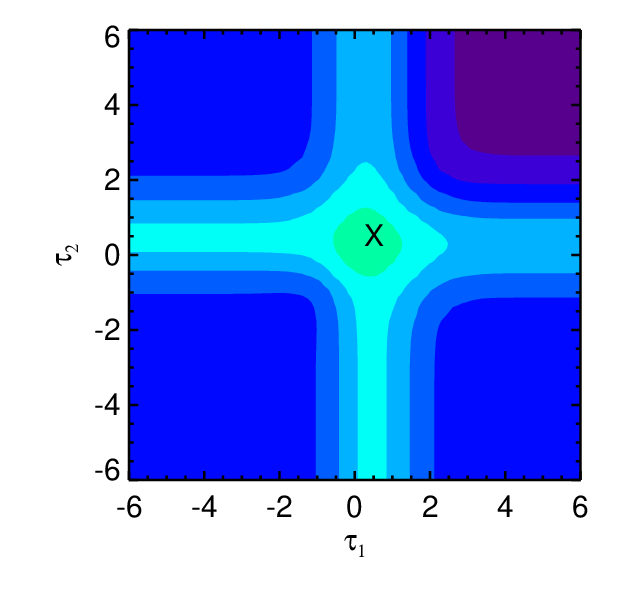} &
\includegraphics[width=1.5in]{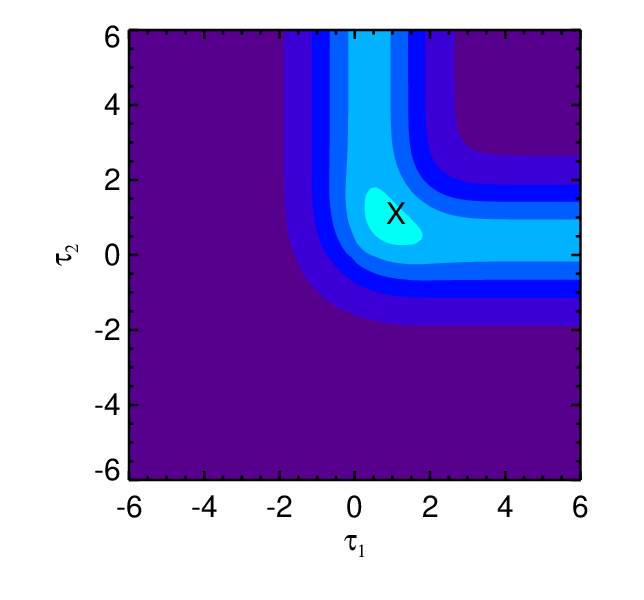}\\
\raisebox{0.8in}{$\sigma_{1}$ = 3.0} &
\includegraphics[width=1.5in]{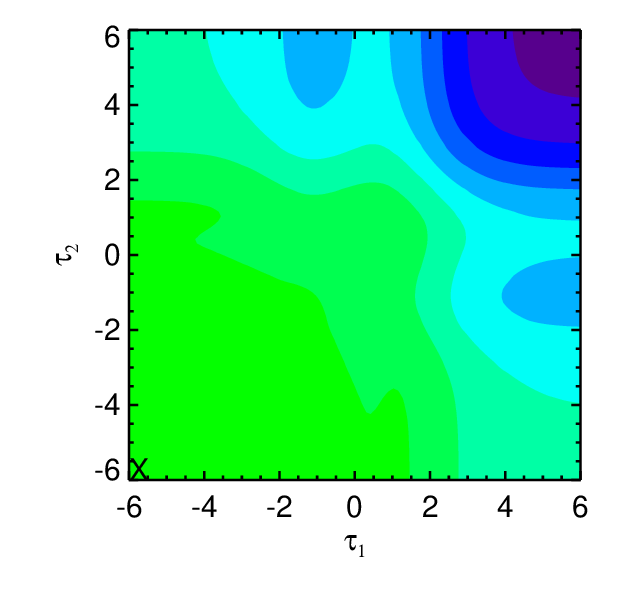} &
\includegraphics[width=1.5in]{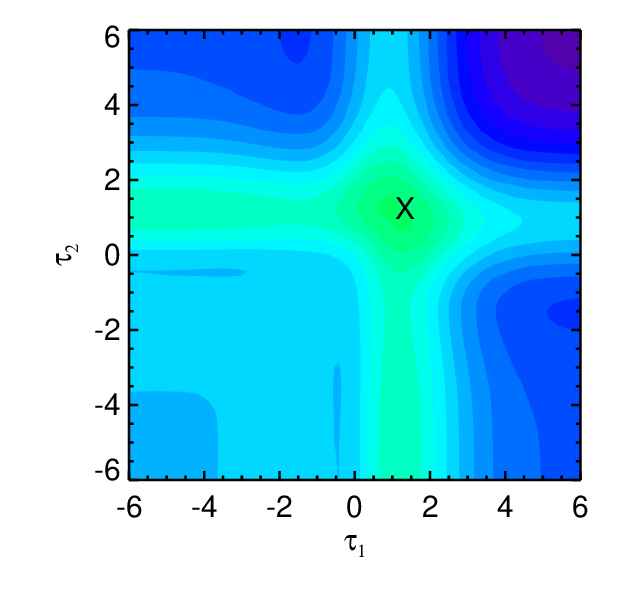} &
\includegraphics[width=1.5in]{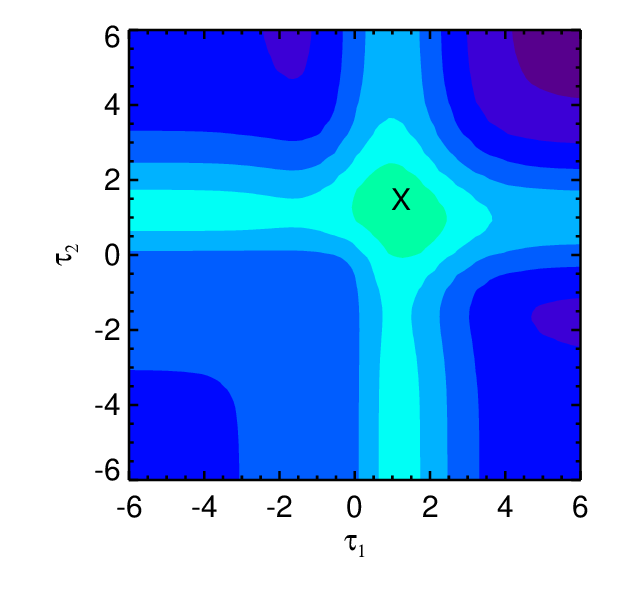} &
\includegraphics[width=1.5in]{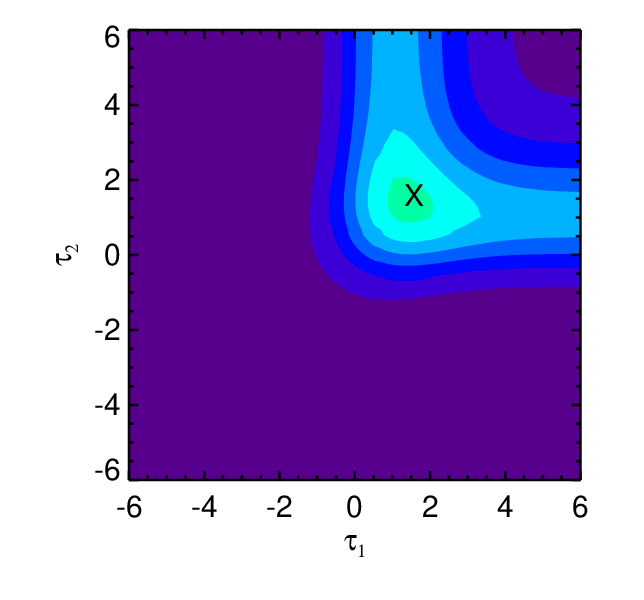}
\end{tabular}
\end{center}
\caption{\label{figF} Results of two-feature trial with x mark showing maximum of ideal observer performance for a given noise variance (shown by sigma) as the data is truncated. The x and y axis shows the range of values for each variable respectively with 0 representing the mid point of the distribution. External noise is shown by $\sigma1$ with normal distribution given by $\sigma1$=1  }
\vspace*{-0.1in}
\end{figure}}

\documentclass[12pt]{article}
\usepackage{graphicx,rangecite,array,outline,amsmath,amsfonts,setspace}
\usepackage{amsmath,amssymb}
\usepackage{bm} 
\newcommand{\vct}[1]{\bm{#1}}
\newcommand{\gvct}[1]{\bm{#1}}

\usepackage{lineno,color}
\usepackage[margin=1.0in]{geometry}
\usepackage{soul} 
\include{mycommands}
\begin{document}
\title{Likelihood ratio for a binary Bayesian classifier under a noise-exclusion model}
\author{Howard~C.~Gifford
\thanks{The author is with the University of Houston, Houston, TX, USA.
E-mail: hgifford@uh.edu}%
}
\date{\small\today}
\maketitle

\newpage
\doublespacing 

\begin{abstract}
We develop a new statistical ideal observer model that performs holistic visual search (or gist) processing in part by placing thresholds on minimum extractable image features. In this model, the ideal observer reduces the number of free parameters thereby shrinking down the system. The applications of this novel framework is in medical image perception (for optimizing imaging systems and algorithms), computer vision, benchmarking performance and enabling feature selection/evaluations. Other applications are in target detection and recognition in defense/security as well as evaluating sensors and detectors.

\end{abstract}


\section{Introduction}\label{sec:introduction}

Assessments of diagnostic image quality can be informative at many stages of technology development in medical imaging, including proof-of-concept research, regulatory approval and in-field application. The appropriate assessments can be stage-dependent, with measures progressing in levels from technical quality (quality control) to societal impact \cite{Fryback:1991wc}. Initial research may center on analyses involving frequency response, whereas subsequent stages may consider comparison with an existing technology that represents an accepted gold standard for a clinically relevant application. One approach related to diagnostic efficiency is assessment uses task-based studies of image quality, which extend technical measures of quality to account for a specific imaging purpose.

An appropriate description of purpose includes what is being imaged, how the images are read and analyzed, and by what measure the analysis is considered correct. For diagnostic applications, the reading can be carried out by either humans or computer models, the latter including statistic models and artificial intelligence (AI).

Task-based assessment thereby enables relevant optimizations of new technology, comparisons of competing technologies, and validation of existing technology for new applications. Taken to a useful extreme, what constitutes ``technology'' generalizes to any inputs to the imaging and analysis process, which includes device design parameters but also characteristics of the imaging subjects, image acquisition protocols, and reader qualifications.

Comparisons of human and model readers are also fundamental to the continued development of task-based assessment itself. Much of the development has come in the context of medical imaging applications and this paper borrows terminology from those works. 
Whether intended to establish ideal performance bounds or mimic humans, these model readers also present constraints. Issues related to task modeling/selection and training requirements can amplify uncertainties, ultimately degrading assessment validity. As general examples, ideal observers (IOs) are often known only approximated. The IOs are also not sensitive to certain classes of invertible (or near-invertible) operations, like some modes of image reconstruction and filtering that are essential for many imaging applications---and may be the subject of analysis. Effects of noise must be implicitly modeled. 

We adopt a mathematical simulation model based on scalar features that are influenced by both external and internal noise sources. The external sources relate to case-to-case stimulus (signal being searched for) variability and data measurement uncertainties. Internal sources include uncertainties from the data processing and non-ideal observer behavior. 

Our main motivations lie in applications of signal detection theory for medical imaging trials. The Bayesian ideal classifier is the statistical discriminant that maximizes the performance of many diagnostic tasks as quantified by the area under the receiver operating characteristic  (ROC) curve. This property establishes a basis for optimizing radiological imaging processes that are degraded by noise stemming from the imaging physics, photon statistics and patient characteristics. A related application evaluates and compares imaging systems through virtual imaging trials that replace radiologists with mathematical models, with the potential to substantially accelerate the clinical adoption of promising technologies \cite{Badano:2018bd}. Some of these models are constructed by equipping ideal classifiers (or simpler-to-devise ideal linear classifiers) with facets of human vision that include spatial frequency filter banks and internal noise that accounts for perceptual and cognitive uncertainties. The ROC rating data generated with these models are essentially weighted sums of image features, with rating variances that reflect both the imaging noise and the internal noise. 
 
We explore new statistical IOs that perform holistic visual-search (or gist) processing in part by placing thresholds on minimum extractable image features. In these models, the IOs reduces the number of free parameters thereby shrinking down the system. The applications of these novel statistical IOs would be in perception models for radiologists decision making in virtual imaging trials for assessment of diagnostic imaging tools. Other areas of impact would be in machine learning, computer vision in benchmarking performance and enabling feature selection/evaluations. These models would also find applications in target detection and recognition in defense/security as well as evaluating sensors and detectors. Aside from medical imaging industrial and non-destructive testing applications would also require robust statistical IOs.


In Sec. 2, we define the task definition, ROC analysis, IOs and how truncation affects the analysis. Section 3 presents mathematical simulation results while Sec. 4 discussion focusing on pros and cons of the truncated IO.


\newpage

\section{Theory}

\subsection{ROC assessment for binary detection tasks}\label{ssec:detect}

The binary task categorizes images as ``negative'' (class $c$ = 0) or ``positive'' ($c$ = 1). In ROC studies, an observer provides for each image a scalar rating statistic $\lambda$ that reflects belief (or confidence) about class membership. Let $p_{\lambda}(\lambda | c)$ represent the class-specific probability distributions for $\lambda$. A diagnostic outcome is rendered by comparing the rating to a decision threshold $t_{\lambda}$. The ratings thus generate a parametric ROC curve showing how the true-positive and false-positive fractions (TPF and FPF) vary with threshold.
Task performance is summarized by the area under the curve \cite{barrett1998a}
\begin{equation}\label{eq:auc}
A_{z} = -\int_{-\infty}^{\infty}\,{\textrm{TPF}}(t_{\lambda})\,\frac{d\, {\textrm{FPF}}(t_{\lambda})}{dt_{\lambda}}\,dt_{\lambda}.
\end{equation}

\subsection{Standard ideal-observer analysis}\label{ssec:roc}

The ideal observer obtains $\lambda$ as a function of image features. A set of $M$ features represented as vector $\vct{f}$ = $(f_{1},\ldots,f_{M})\in\mathbb{R}^{M}$ is available. The class-specific joint feature distributions are denoted as $p(\vct{f} | c)$ and the observer maximizes $A_{z}$ by computing ratings according to the log-likelihood ratio \cite{geisler:2003vc} 
\begin{equation}\label{eq:idealstat}
\lambda(\vct{f}) = \ln\frac{p(\vct{f} | c=1)}{p(\vct{f} | c=0)}.
\end{equation} 
Feature variability in this ideal case is imparted by external sources---class membership and measurement noise in the data collection. We write $p(\vct{f} | c)$ = $p_{\mathrm{ext}}(\vct{f} | c)$ to emphasize the origins of this noise in the absence of other sources. 


Observer inefficiencies in dealing with the measured features qualify as internal noise that impacts decision making.  These inefficiencies are emblematic of suboptimal observers; our objective is to investigate optimal procedures for handling inefficiencies that are recognized by the observer. Two forms of inefficiency are considered. One form occurs when log-likelihood ratios are computed on the basis of approximate conditional distributions $\hat{p}_{\mathrm{ext}}(\vct{f} | c)$. The approximations may be introduced by uncertainties in the parameter estimation for $p(\vct{f} | c)$ or by omission of specific distribution components. A common instance of the latter is applying a nonprewhitening matched filter in tasks with correlated features. 

The second inefficiency is postmeasurement processing noise, which is treated herein as additive and independent with regard to features and class. In the absence of data truncation, our observer makes use of the multivariate probability distributions 
\begin{equation}\label{eq:notrunc}
p(\vct{f} | c) = p_{\mathrm{ext}}(\vct{f} | c)\ast\,p_{\mathrm{int}}(\vct{f})
\end{equation}
obtained by the $M$-dimensional convolution of the measurement and processing distributions. The internal-noise distribution $p_{\mathrm{int}}$ is considered separable based on these assumptions.

\subsection{Truncated feature distributions}\label{ssec:features}

Data truncation introduces explicit feature extraction into the ideal observer, separating the two noise processes so that the internal noise acts to penalize the use of noninformative data. In this development, we consider lefthand (or lower) truncation as implemented with unit-height step functions $\mathrm{step}(f_{i}-\tau_{i})$, which preserve feature values $f_{i} \ge \tau_{i}$ for $i=1,\ldots,M$. Righthand and dual truncation are other options for modifying the relative emphasis on true-positives and true-negatives in the decision making. The set of lefthand thresholds is grouped in the vector $\gvct{\tau} = (\tau_{1},\ldots,\tau_{M})$. A typical computational objective would be to optimize this vector based on the ROC area; it is not clear that human observers could accomplish the calculations given the combinatorical dimensions involved. The example simulations in this paper simply scan the thresholds in one and two dimensions.

The thresholds divide the feature space 
into 
$2^{M}$ disjoint regions, each defining a unique subset of extracted features for the observer. We associate each subset with an $M$-element indicator vector $\gvct{\alpha}$ containing values $\alpha_{i}$ that are respectively 1 and 0 for extracted and rejected feature indices. The empty set, corresponding to the event that all features are rejected and a given image goes unrated, has $\alpha_{i}$ = 0 for all $i$, to be denoted as $\gvct{\alpha}$ = {\bf{0}}. The instance of complete extraction has $\gvct{\alpha}$ = $\vct{1}$. The region associated with a given $\gvct{\alpha}$ under lefthand truncation has the feature support $\gvct{\chi}_{\alpha}(\gvct{\tau})$ = \{$\vct{f}$\,:\,$S_{\alpha}(\vct{f},\gvct{\tau})$ = 1\} as specified by the composite step function
\begin{equation}\label{eq:alphatrunc}
S_{\alpha}(\vct{f},\gvct{\tau})=\prod_{i=1}^{M}\left[1-\mathrm{step}(f_{i}-\tau_{i})\right]^{1-\alpha_{i}}\left[\mathrm{step}(f_{i}-\tau_{i})\right]^{\alpha_{i}}.
\end{equation}
The complement set \{$\vct{f}$\,:\,$S_{\alpha}(\vct{f},\gvct{\tau})$ = 0\} defines the support for $\gvct{\alpha}$ under righthand truncation. 

With this support notation, the $\gvct{\alpha}$-extraction probability---the probability of extracting a particular subset of features---can be expressed as
\begin{equation}\label{eq:alphaprob}
Pr_{\mathrm{ext},\tau}(\gvct{\alpha}|c)
=\int_{\gvct{\chi}_{\alpha}(\gvct{\tau})}\,p_{\mathrm{ext}}(\vct{f}|c)\,d\vct{f}.
\end{equation}
With independent features, these probabilities are products of rejection and complementary acceptance probabilities for the individual features. The probability of rejecting all the features for an image, also calculable as
\begin{equation}\label{eq:alphaprob0}
Pr_{\mathrm{ext},\tau}(\gvct{\alpha} = \vct{0}|c)=1 - \sum_{\gvct{\alpha}\ne\vct{0}}\,Pr_{\mathrm{ext},\tau}(\gvct{\alpha}|c),
\end{equation}
will be of particular consideration in our ROC analysis. 

Describing the feature distribution for nonzero $\gvct{\alpha}$ calls for separating the elements of $\vct{f}$ into a pair of variable-length vectors $\left(\vct{e},\vct{e}'\right)$, where $\vct{e}$ contains the $m$ extracted elements ($f_{i}$ such that $\alpha_{i}$ = 1) and the complement $\vct{e}'$ contains the $M$ - $m$ rejected elements ($f_{i}$ for which $\alpha_{i}$ = 0). With full feature information (i.e., $m$ = $M$), the distribution is
\begin{equation}\label{eq:alphamarg}
p_{\mathrm{ext},\tau}(\vct{e}|\gvct{\alpha},c)=\frac{1}{Pr_{\mathrm{ext},\tau}(\gvct{\alpha}|c)}\,p_{\mathrm{ext}}(\vct{f}|c)\ S_{\alpha}(\vct{f},\gvct{\tau}).
\end{equation}
Otherwise, the distribution is obtained from the marginal calculation
\begin{equation}\label{eq:alphamarg2}
p_{\mathrm{ext},\tau}(\vct{e}|\gvct{\alpha},c)=\frac{1}{P_{\mathrm{ext},\tau}(\gvct{\alpha}|c)}\,\int_{\mathbb{R}^{M-m}}p_{\mathrm{ext}}(\vct{f}|c)\ S_{\alpha}(\vct{f},\gvct{\tau})\,d\vct{e}'.
\end{equation}
We refer to $p_{\mathrm{ext},\tau}(\vct{e}|\gvct{\alpha},c)$ as the truncated external source distribution. Note there is no need for a distribution component for $\gvct{\alpha}$ = $\vct{0}$ as the impact of total feature rejection will be addressed with a separate term in the ROC area formula.

The final step to our data model is insertion of internal noise via convolution in the manner of Eq.~\ref{eq:notrunc}, so that
\begin{equation}\label{eq:trunc}
p_{\tau}(\vct{e}|\gvct{\alpha},c)= p_{\mathrm{ext},\tau}(\vct{e}|\gvct{\alpha},c)\ast\,p_{\mathrm{int}}(\vct{e}),
\end{equation}
with the caveat that $p_{\mathrm{ext},\tau}(\vct{e}|\gvct{\alpha},c)$ is a function of $m$ features and the convolution applies only to that feature subset.

\subsection{ROC analysis with truncated data}\label{ssec:tradeoffs}

The general tradeoff in optimizing the thresholds is evident: controlling the onset of internal noise while also discarding potentially useful data. This negative effect of truncation is partially mitigated by the multivariate data, in that a rating can be obtained with but a single retained feature value. However, even moderate truncation results in substantial ROC data loss; three independently distributed features with identical acceptance probabilities of $\frac{1}{2}$ cause one image in eight to be unrated. How unrated cases are treated in the ROC analysis greatly affects the optimization \cite{Karbaschi:2018gk}.


An incomplete analysis excludes unrated cases from the true-positive and false-positive tallies while preserving them for the denominators of the TPF and FPF calculations. With lefthand truncation, this leads to a partial ROC curve that terminates at the coordinates
\begin{equation}\label{eq:rocpartial}
(\mathrm{TPF},\mathrm{FPF}) = (1 - Pr_{\mathrm{ext},\tau}(\gvct{\alpha}=\vct{0}|c=1),1 - Pr_{\mathrm{ext},\tau}(\gvct{\alpha}=\vct{0}|c=0)).
\end{equation}
As related in Sec.~\ref{ssec:features}, the $Pr_{\mathrm{ext},\tau}$ terms are the probabilities of full feature rejection for the two image classes. For purposes of optimizing $\gvct{\tau}$, the considerable negative impact on $A_{z}$ from partial curves (see Fig.~\ref{figA}) often swamps the effects of internal noise, in which case the optimal solution is to complete the ROC curve by reducing the truncation. 

A meaningful optimization completes the curve. How this can be done under lefthand truncation is evident in the context of forced-choice studies. Observers are tasked with identifying the positive case from an image pair and $A_{z}$ is the probability of success. Area $A_{z}^{(1)}$ of the shaded region in Fig.~\ref{figA} represents the joint probability that the pair are rated and correctly labeled. We consider this the analysis component of observer performance. The region delineated by the horizontal dashed extension in Fig.~\ref{figA}, with area
\begin{equation}\label{eq:del_az2}
A_{z}^{(2)}=\left[1 - Pr_{\mathrm{ext},\tau}(\gvct{\alpha}=\vct{0}|c=1)\right]\,Pr_{\mathrm{ext},\tau}(\gvct{\alpha}=\vct{0}|c=0),
\end{equation}
results from comparing unrated negative cases against rated positive images. From an ROC perspective, this comes about with a modification in FPF given by  
\begin{equation}\label{eq:del_fpf}
\Delta FPF=Pr_{\mathrm{ext},\tau}(\gvct{\alpha}=\vct{0}|c=0)\,\delta(t_{\lambda}+\infty),
\end{equation}
accorded by preemptorily assigning (through the Dirac $\delta$-function) the minimum rating $\lambda$ = $-\infty$ to the unrated negative images. The association of unrated cases with de facto negative classification leads us to refer to $A_{z}^{(2)}$ as the gist component of observer performance. 

\figA

Forced-choice comparisons of unrated positive images against rated negative images do not augment $A_{z}$. However, a third performance contribution comes from observer guessing with pairs of unrated images, with even odds of success in the binary task. This guessing component completes the ROC curve with the dotted straight-line segment to the point (1,1) as show in Fig.~\ref{figA}. 
The triangle area
\begin{equation}\label{eq:del_az3}
A_{z}^{(3)}=\frac{1}{2}\,Pr_{\mathrm{ext},\tau}(\gvct{\alpha}=\vct{0}|c=1)\,Pr_{\mathrm{ext},\tau}(\gvct{\alpha}=\vct{0}|c=0)
\end{equation}
is also associated with the FPF increment of Eq.~\ref{eq:del_fpf} in combination with the TPF increment
\begin{equation}\label{eq:del_tpf}
\Delta TPF=Pr_{\mathrm{ext},\tau}(\gvct{\alpha}=\vct{0}|c=1)\,\delta(t_{\lambda}+\infty).
\end{equation}

The gist and guessing components $A_{z}^{(2)}$ and $A_{z}^{(3)}$ are thus determined by the external-source probability distributions and $\gvct{\tau}$. There is no influence from the underlying observer model or internal noise, although these are deciding factors in how truncation affects the analysis component of performance. All else being equal, optimizations that include the guessing component will tend towards greater data truncation compared to optimizations that neglect the guessing component. Optimizations in which data truncation leads to increased $A_{z}$ can be interpreted as improving performance with relatively simple cases (compared to no truncation) at the expense of reduced performance with more ambivalent cases.

\subsection{Ideal observers with truncated data}\label{ssec:observers}

Our ideal observer computes ratings with the log-likelihood ratio that is appropriate for the data from a given image, generalizing Eq.~\ref{eq:idealstat} to
\begin{equation}\label{eq:idealstat2}
\lambda_{\alpha,\tau}(\vct{e}) = \ln\frac{p_{\tau}(\vct{e}|\gvct{\alpha},c=1)}{p_{\tau}(\vct{e}|\gvct{\alpha},c=0)}.
\end{equation}
The class-conditional rating distributions derived from Eq.~\ref{eq:idealstat2} are represented as $p_{\lambda,\alpha,\tau}(\lambda | c)$. Overall, this model generates ratings from the mixture distributions
\begin{equation}\label{eq:mixture}
p_{\lambda,\tau}(\lambda | c) = \sum_{\gvct{\alpha}\ne\vct{0}}\,Pr_{\mathrm{ext},\tau}(\gvct{\alpha}|c)\,p_{\lambda,\alpha,\tau}(\lambda | c).
\end{equation}
Relevant ROC quantities computed from the rated images alone are 
\begin{align}
{\mathrm{TPF}}(t_{\lambda}) = & \sum_{\gvct{\alpha}\ne\vct{0}}\,Pr_{\mathrm{ext},\tau}(\gvct{\alpha}|c=1)\,\int_{t_{\lambda}}^{\infty}\,p_{\lambda,\alpha,\tau}(\lambda |c=1)\,d\lambda\label{eq:llr4}\\
\frac{d\, {\mathrm{FPF}}(t_{\lambda})}{dt_{\lambda}} = & \,\sum_{\gvct{\alpha}\ne\vct{0}}\,Pr_{\mathrm{ext},\tau}(\gvct{\alpha}|c=0)\,p_{\lambda,\alpha,\tau}(t_{\lambda} |c=0).\label{eq:llr5}
\end{align}
Following Eq.~\ref{eq:auc} and including the area components $A_{z}^{(2)}$ and $A_{z}^{(3)}$ related to unrated images, the ROC performance is
\begin{equation}\label{eq:subsetstat}
A_{z}= \,A_{z}^{(1)}+\left[1 - \frac{1}{2}Pr_{\mathrm{ext},\tau}(\gvct{\alpha}=\vct{0}|c=1)\right]\,Pr_{\mathrm{ext},\tau}(\gvct{\alpha}=\vct{0}|c=0)
\end{equation}
with the rating-derived component
\begin{equation}\label{eq:subsetstat2}
A_{z}^{(1)}=\sum_{\gvct{\alpha},\tilde{\gvct{\alpha}}\ne\vct{0}}\,Pr_{\mathrm{ext},\tau}(\gvct{\alpha}|c=1)\,Pr_{\mathrm{ext},\tau}(\tilde{\gvct{\alpha}}|c=0)\,A_{z,\alpha,\tilde{\alpha},\tau}^{(1)},
\end{equation}
as the weighted sum of cross-$\gvct{\alpha}$ area contributions
\begin{equation}\label{eq:subsetstat3}
A_{z,\alpha,\tilde{\alpha},\tau}^{(1)}=-\int_{-\infty}^{\infty}\,p_{\lambda,\tilde{\alpha},\tau}(t_{\lambda}|c=0)\,\int_{t_{\lambda}}^{\infty}\,p_{\lambda,\alpha,\tau}(\lambda |c=1)\,{d\lambda}\,{dt_\lambda}.
\end{equation}

Using untruncated ratings without internal noise, the ideal observer generates a nonconcave ROC curve that places an upper bound on curves containing linear segments (see Fig.~\ref{figB}). With no truncation and a fixed internal noise magnitude, the observer provides a lower bound on ROC curves obtained using truncation with the same internal-noise magnitude. The optimal curve under truncation vector ${\gvct{\tau}}$ intersects the upper-bound curve at the threshold operating point given by Eq.~\ref{eq:rocpartial}, indicating that ideal performance is achievable if decisions can be made with minimally processed extracted features so as to avoid subsequent sources of internal noise. The intersection point is a function of ${\gvct{\tau}}$, and in the simulations of Sec.~\ref{ssec:sims}, the optimal ${\gvct{\tau}}$ is identified by maximizing $A_{z}$.  

\figB

\subsection{Ideal observers with high internal noise}\label{ssec:observers-high-noise}

In the limit of high internal noise for any observer model, component $A_{z,\alpha,\tilde{\alpha},\tau}^{(1)}$ evaluates to $\frac{1}{2}$, and the area under the curve as a function of $\gvct{\tau}$ for the ideal observer with truncated ratings is
\begin{equation}\label{eq:auc-asympt}
A_{z}=\frac{1}{2}\left[1+Pr_{\mathrm{ext},\tau}(\gvct{\alpha}=\vct{0}|c=0)-Pr_{\mathrm{ext},\tau}(\gvct{\alpha}=\vct{0}|c=1)\right].
\end{equation}
The next section examines optimization based on this asymptotic result as well as other properties of the truncated ideal observer with the number of features $M$ $\in$ \{1,2\}.

\section{Computer simulations}\label{ssec:sims}

We consider normally distributed features, adopting the notation 
$p_{\mathrm{ext}}(\vct{f}|c) = {\mathcal{N}}(\gvct{\mu}_{c},\gvct{\Sigma}_{c})$
for the external effects, with $\gvct{\mu}_{c}$ and $\gvct{\Sigma}_{c}$ respectively the mean feature vector and $M\times M$ covariance matrix for class $c$. The class mean of feature $f_{i}$ is $\mu_{c,i}$ and the class variance is $\sigma_{c,i}^2$. Both independent and correlated feature models are tested. All trials treat normally distributed and zero-mean internal noise, so that    
$p_{\mathrm{int}}(\vct{f}) = {\mathcal{N}}(\vct{0},\sigma^{2}\vct{I})$,
with variance $\sigma^{2}$ and $\vct{I}$ the $M$-dimensional identity matrix. This specific instance of the thresholding observer model presents $M$ feature thresholds, 2$M$ feature means, 2$M+1$ feature variances, and $M^{2}-M$ feature covariance parameters. With independent features, there are no covariances. 


\subsection{Single-feature trials}\label{sec:results_}

Some basic properties of probability distribution shaping are evident with $M$ = 1. We assign the standard external-source distribution ${\mathcal{N}}(0,1)$ to the negative class and test positive-class variations ${\mathcal{N}}(0.75,\sigma_{1}^{2})$ such that $\sigma_{1}^{2}$ is less than, equal to, or greater than 1. Without truncation, the ideal observer is a linear function in the homoskedastic case and quadratic otherwise. Invoking truncation renders the homoskedastic model piecewise linear while adding higher-order terms to the quadratic models. Complete information for the ideal observer is contained within the processed distributions of Eq.~\ref{eq:trunc}, which for single or independent normal features have the closed-form expression
\begin{equation}\label{eq:convolpdf1}
p_{\tau}(\vct{e}|\gvct{\alpha},c)=\prod_{\substack{i=1\\ \alpha_{i}=1}}^{M}\,{\mathcal{N}}(\mu_{c,i},\sigma_{c,i}^{2}+\sigma^{2})\Phi(\frac{f_{i}-a_{c,i}}{b_{c,i}})\,/\Phi(\frac{\mu_{c,i}-\tau_{i}}{\sigma_{c,i}}),
\end{equation}
with $\Phi(\cdot)$ the standard normal cumulative distribution function (CDF) and parameters
\begin{align}
a_{c,i}=&\frac{1}{\sigma_{c,i}^{2}}\left[\tau_{i}\left(\sigma_{c,i}^{2}+\sigma^{2}\right)-\sigma^{2}\mu_{c,i}\right]\\
b_{c,i}=&\frac{\sigma}{\sigma_{c,i}}\sqrt{\sigma_{c,i}^{2}+\sigma^{2}}.
\end{align}
As the internal noise is reduced (i.e., $\sigma$ $\rightarrow$ 0), the CDF approaches the limit of a step function with discontinuity at $f_{i}$ = $\tau_{i}$.

The upper row of graphs in Fig.~\ref{figE} confirms the optimality of untruncated performance with $\sigma$ = 0. Unregularized performance decreases monotonically with $\tau$ as shown by the red plot lines. The degree of extended stability under the regularized model (black lines) is dictated by the effectiveness of the gist process (blue lines) at solely truncating negative cases, thus forestalling the guessing (green lines) associated with increased positive-case truncation. For $\sigma_{1}$ = 3, the instability at lower $\tau$ results from preferential truncation of positive cases (which is not fully represented in the guessing component). A subsequent result is the appearance of a secondary performance peak for values of $\tau$ between the class means. 

The significance of this thresholding peak emerges with increased internal noise for all $\sigma_{1}$ as evidenced in the lower row of plots in Fig.~\ref{figE}. The threshold advantage may appear to be greater for smaller values of $\sigma_{1}$ due to more-effective gist processing, but under at least one relevant measure the effect is consistent across $\sigma_{1}$. We define minimum truncation efficiency as the ratio of truncation performance for $\sigma$ to the performance of the untruncated ideal observer. The effect is most striking at high internal noise, representing greatest possible benefit of PDS, to the extent that optimal observer performance under extreme internal noise is considerably better than the 0.5 obtained by the ideal observer with added internal noise. Optimal performance exceeds chance (50\%) even in the absence of informative ratings (i.e., at infinite variance). 

\figE


\subsection{Two-feature trials}\label{sec:results__}

When $M$ is greater than 1, interactions between features can affect observer performance. Studies with two independent features were carried out to illustrate this. 

We assign the standard external-source distribution ${\mathcal{N}}(0,1)$ to the negative class and test positive-class variations ${\mathcal{N}}(0.75,\sigma_{1}^{2})$ such that $\sigma_{1}^{2}$ is less than, equal to, or greater than 1. Without truncation, the ideal observer is a linear function in the homoskedastic case and quadratic otherwise. Invoking truncation renders the homoskedastic model piecewise linear while adding higher-order terms to the quadratic models. 

Complete information for the ideal observer is contained within the processed distributions of Eq.~\ref{eq:trunc}, which for single or independent normal features have the closed-form expression
\begin{equation}\label{eq:convolpdf1}
p_{\tau}(\vct{e}|\gvct{\alpha},c)=\prod_{\substack{i=1\\ \alpha_{i}=1}}^{M}\,{\mathcal{N}}(\mu_{c,i},\sigma_{c,i}^{2}+\sigma^{2})\Phi(\frac{f_{i}-a_{c,i}}{b_{c,i}})\,/\Phi(\frac{\mu_{c,i}-\tau_{i}}{\sigma_{c,i}}),
\end{equation}
with $\Phi(\cdot)$ the standard normal cumulative distribution function (CDF) and parameters
\begin{align}
a_{c,i}=&\frac{1}{\sigma_{c,i}^{2}}\left[\tau_{i}\left(\sigma_{c,i}^{2}+\sigma^{2}\right)-\sigma^{2}\mu_{c,i}\right]\\
b_{c,i}=&\frac{\sigma}{\sigma_{c,i}}\sqrt{\sigma_{c,i}^{2}+\sigma^{2}}.
\end{align}


%
%

\figF



With the additional assumption that the features are identically distributed, a single threshold is required regardless of $M$. The effect of threshold, processing noise variance and number of features on IO performance is shown in Fig.~\ref{figF}. The figure shows local maxima at truncated data with $\vct{\tau}$ near the dual-class means.

The IO is sensitive to the truncation of features, as opposed to variations in feature values. Observers with different levels of internal noise can perform similarly by using well-adjusted thresholds, although the penalties for misadjusted thresholds are amplified with internal noise. 

\newpage

%
%
%
%
%
%

\section{Discussion}\label{sec:discussion}

We have introduced an ideal observer for a noise-exclusion model. Data at a given region are truncated by a fixed lower threshold. Analysis, gist and guessing contributions make up the AUC in the binary ideal observer for truncated data. In the canonical two-choice model, the analysis contribution is from comparing rated normal and abnormal cases, while the gist component comes from comparisons of one rated case and one unrated case. The guessing component comes from comparing unrated normal and abnormal cases. 

With a low amount of noise, the analysis component is the major contributor, and AUC is highest for low truncation. As noise ramps up, the gist process becomes the leading component. Guessing is important for excessive noise. 

When the IO is considered as a suboptimal observer compared to the traditional IO lacking internal noise, it is of interest to learn how much additive internal noise is required for traditional IO to match the suboptimal IO.

\section{Acknowledgments}
This work was supported by the National Institute for Biomedical Imaging and Bioengineering 
under grant number R01EB032416. The contents are solely the responsibility of the author and do not necessarily represent official NIBIB views.
\bibliographystyle{ieeetr}
\bibliography{totalrefs,papers2_biblib}

\begin{thebibliography}{1}

\bibitem{Fryback:1991wc}
D.~G. Fryback and J.~R. Thornbury, ``{The efficacy of diagnostic imaging.},''
  {\em Medical Decision Making}, vol.~11, pp.~88--94, Apr. 1991.

\bibitem{Badano:2018bd}
A.~Badano, C.~G. Graff, A.~Badal, D.~Sharma, R.~Zeng, F.~W. Samuelson, S.~J.
  Glick, and K.~J. Myers, ``{Evaluation of Digital Breast Tomosynthesis as
  Replacement of Full-Field Digital Mammography Using an In Silico Imaging
  Trial.},'' {\em JAMA network open}, vol.~1, p.~e185474, Nov. 2018.

\bibitem{barrett1998a}
H.~H. Barrett, C.~K. Abbey, and E.~Clarkson, ``Objective assessment of image
  quality. {III}. {ROC} metrics, ideal observers, and likelihood-generating
  functions,'' {\em J. Opt. Soc. Am. A}, vol.~15, pp.~1520--1535, 1998.

\bibitem{geisler:2003vc}
W.~S. Geisler, {\em {Ideal observer analysis}}.
\newblock 2003.

\bibitem{Karbaschi:2018gk}
Z.~Karbaschi and H.~C. Gifford, ``{Assessing CT acquisition parameters with
  visual-search model observers.},'' {\em Journal of Medical Imaging}, vol.~5,
  no.~2, p.~025501, 2018.

\end{thebibliography}
\end{document}